\documentclass[twoside,a4paper,12pt]{article}
\usepackage[sc]{mathpazo} 
\usepackage[T1]{fontenc} 
\usepackage{microtype} 

\usepackage[english]{babel}

\usepackage[hang, small,labelfont=bf,up,textfont=it,up]{caption} 
\usepackage{booktabs} 
\usepackage{graphicx}
\usepackage{amsmath}
\usepackage{comment}
\usepackage{multirow}
\usepackage{color}
\usepackage{soul}

\makeatletter
\renewcommand*{\@fnsymbol}[1]{\ensuremath{\ifcase#1\or \dagger\or \ddagger\or * \or
   \mathsection\or \mathparagraph\or \|\or **\or \dagger\dagger
   \or \ddagger\ddagger \else\@ctrerr\fi}}
\makeatother

\usepackage{enumitem}
\setlist[itemize]{noitemsep} 

\usepackage{abstract}

\usepackage{titlesec} 
\renewcommand\thesection{\Roman{section}}
\titleformat{\section}[block]{\large\scshape\centering}{\thesection.}{1em}{} 
\titleformat{\subsection}[block]{\large}{\thesubsection.}{1em}{}

\usepackage{fancyhdr} 
\usepackage{titling} 

\usepackage{hyperref} 
\usepackage[capitalize,noabbrev]{cleveref}
\usepackage{fullpage}
\usepackage{tikz}
\usepackage[edges]{forest}
\usepackage{bm}
\usepackage{float}
\usetikzlibrary{arrows.meta, positioning, backgrounds}
\usepackage{graphicx}
\usepackage{lineno}

\forestset{
    .style={
        for tree={
            base=bottom,
            child anchor=north,
	   parent anchor=south,
            align=center,
            s sep+=1cm,
	   inner sep=0pt,
    straight edge/.style={
        edge path={\noexpand\path[\forestoption{edge},thick,-{Latex}] 
        (!u.parent anchor) -- (.child anchor);}
    },
    if n children={0}
        {tier=word, draw, thick, rectangle}
        {draw, diamond, thick, aspect=2},
    if n=1{%
        edge path={\noexpand\path[\forestoption{edge},thick,-{Latex}] 
        (!u.parent anchor) -| (.child anchor) node[pos=.2, above] {Y};}
        }{
        edge path={\noexpand\path[\forestoption{edge},thick,-{Latex}] 
        (!u.parent anchor) -| (.child anchor) node[pos=.2, above] {N};}
        }
        }
    }
}

\usepackage[numbers,sort&compress,super]{natbib}
\setcitestyle{authoryear,round,citesep={;},aysep={,},yysep={;}}

\definecolor{myblue}{RGB}{143, 186, 212}
\definecolor{myred}{RGB}{200, 121, 137}
\definecolor{mypurple}{RGB}{178, 169, 200}
\definecolor{myorange}{RGB}{183, 122, 73}

\pretitle{\begin{center}\Huge\bfseries} 
\posttitle{\end{center}} 
\title{Modelling customer lifetime-value in the retail banking industry} 
\author{%
\textsc{Greig Cowan}$^1$ 
\hspace{20pt} 
\textsc{Salvatore Mercuri}$^1$
\hspace{20pt}\textsc{Raad Khraishi}\noindent$^{1,}$ $^{2\enspace\ignorespaces}$ \\[1ex]
\small $^1$\ignorespaces Data Science and Innovation, NatWest Group, London, United Kingdom\thanks{Correspondence to: \href{mailto:greig.cowan@natwest.com}{greig.cowan@natwest.com}} \\[1ex]
\small $^2$\ignorespaces Institute of Finance and Technology, UCL, London, United Kingdom\thanks{Correspondence to: \href{mailto:raad.khraishi@ucl.ac.uk}{raad.khraishi@ucl.ac.uk}}
}
\date{} 
\renewcommand{\maketitlehookd}{%
\begin{abstract}
\noindent 
Understanding customer lifetime value is key to nurturing long-term customer relationships, however, estimating it is far from straightforward. 
In the retail banking industry, commonly used approaches rely on simple heuristics and do not take advantage of the high predictive ability of modern machine learning techniques.
We present a general framework for modelling customer lifetime value which may be applied to industries with long-lasting contractual and product-centric customer relationships, of which retail banking is an example. 
This framework is novel in facilitating CLV predictions over arbitrary time horizons and product-based propensity models.
We also detail an implementation of this model which is currently in production at a large UK lender. 
In testing, we estimate an 43\% improvement in out-of-time CLV prediction error relative to a popular baseline approach.
Propensity models derived from our CLV model have been used to support customer contact marketing campaigns. In testing, we saw that the top 10\% of customers ranked by their propensity to take up investment products were 3.2 times more likely to take up an investment product in the next year than a customer chosen at random.
\end{abstract}
}

\begin{document}

\maketitle


\section{Introduction\label{sec:intro}}

Modern marketing strategies are a result of a process that began in the 1980s, with a shift from analysing individual transactions towards focusing on the holistic buyer-seller relationship \citep{Dwyer1987}. This shift was  empirically shown to drive profitability \citep{Reichheld1990, Srivastava1999, Gupta2006CustomerMA}; for example, \cite{Reichheld1990} showed that, across over 100 tested credit card companies, the profit generated by customers increased the longer a customer stayed with the company. 
This shift has continued up to the present period in which it is widely accepted that the primary purpose of a business is to nurture symbiotic customer relationships \citep{Ryals2005}. 
However, customer relationships and their long-term value are individual in nature.
As a result, it has become increasingly important to evaluate, at a customer level, the Customer Lifetime Value (CLV), which was first defined by \cite{kotler1974} as the ``present value of the future profit stream expected over a given time horizon of transacting with the customer''.

Early developments in modelling CLV focused on conceptual frameworks that addressed practical challenges on an industry-agnostic level \citep{berger1998, jain2002, rust2004, Gupta2006, pfeifer2000modeling, Blattberg2008, venkatesan2007optimal}. 
With a few exceptions \citep{keane1995, berger2003} in this earlier period, there was little focus on industry-specific applications of CLV models. This is surprising since customer behaviour and the buyer-seller relationship is heavily dependent on the type of industry to which the seller belongs \citep{Haenlein2007}. 
In recent years, as CLV modelling became an integral component in marketing strategies, there has been a shift to industry-specific approaches, such as for car repair and maintenance \citep{cheng2012customer}, telecom \citep{Flordahl2013}, e-commerce \citep{Paauwe2007}, and financial services \citep{Ramn2014AMO}. 
With respect to the retail banking industry, the first notable work in this area was by \cite{Haenlein2007}, which modelled the customer relationship as a first-order Markov decision process. 
The application of this model, particularly in the context of marketing campaigns, was further applied and validated by \cite{ekinci2014analysis} and in \cite{tomas}. 
A logistic regression model to predict one-year ahead, an alternative to the more dominant first-order Markov process approach, was developed by \citet{ekinci2014customer2}. 
More recently, \citet{Mosaddegh2021} developed a Recency-Frequency-Monetary Value (RFM) model to predict CLV in the banking industry, however this does not take into account any customer demographics. 
\cite{MendezSuarez2021} utilised real option theory to predict CLV in order to explain why even unprofitable customers should be retained in the banking industry.

Machine learning models are increasingly applied for a broad range of modelling purposes, in part due to their high predictive performances. 
This is true also of CLV modelling, with recent work of \citet{chamberlain2017customer} using neural networks to generate customer embeddings as input features to ensemble regressors to predict CLV for an online fashion retailer. 
However, the explainability of a CLV model is often a priority due to its role in supporting marketing campaigns, and the black-box decisioning of some machine learning models has led to a lower-than-expected prevalence in CLV modelling.

In the current paper, we demonstrate a methodology for modelling CLV at a UK-based retail bank, which focuses on the following three pillars:
\begin{itemize}
	\item Understanding customer needs through personalised CLV modelling. 
	\item Addressing these needs through a variety of contact and engagement strategies. 
	\item Iterating engagement and modelling strategies according to testing and feedback. 
\end{itemize}
Our modelling methodology combines that of \cite{Haenlein2007} with state-of-the-art machine learning components that are developed with minimal black-box decisioning in order to support product-based marketing campaigns. 
In addition, our approach is novel in allowing for predictions over longer time horizons, through a multi-year simulator that can predict CLV over arbitrary time horizons, as well as incorporating multiple products that may be used to support various product-based marketing campaigns.

In \Cref{sec:model} we describe our modelling methodology along with key assumptions. 
Note that this methodology may be applied to other industries outside of retail banking. 
In \Cref{sec:application} we detail an application of this modelling methodology at a retail bank, discussing limitations, data, and aspects of the modelling that are specific to this use case. 
Moreover, we validate the model on both in-time and out-of-time test sets.
Finally, we give some concluding remarks in \Cref{sec:conclusion}.

\section{Methodology\label{sec:model}}

In this section, we describe the modelling framework at a high level. 
This framework is suitable for any industry that consists primarily of long-lasting contractual customer relationships, especially those that involve a relatively small number of products that generate a large amount of value. 
In \Cref{sec:application}, we describe a concrete application of this model in the retail banking industry.

\subsection{Definition of CLV\label{sec:clv_def}}

We define a customer's value (CV) for a given time period in terms of the value (to the company) of their product holdings in that time period, and their lifetime value (CLV) as the sum of the values for each period in the given time horizon. 
Formally, the \emph{value} of a product $p$ held by customer $c$ in period $t$ is given by:
\begin{equation}\label{eq:ev}
	V_t^{c, p} := R_t^{c, p} - C_t^{c, p},
\end{equation}
where, for customer $c$'s product $p$, $R_t^{c, p}$ denotes the product's \emph{revenue} (e.g., interest income) and $C_t^{c, p}$ denotes the total \emph{cost} of the product to the company (which could include, e.g.,  the cost of default).

The \emph{customer value} of customer $c$ in time period $t$ is then the sum
\begin{equation}\label{eq:cv}
	CV_t^c := \sum_p V_t^{c, p},
\end{equation}
which ranges over all products $p$ held by customer $c$. Finally, the \emph{customer lifetime value} of that customer over a time horizon of $T$ is given by
\begin{equation}\label{eq:clv}
	CLV_T^c := \sum_{t=1}^T \frac{CV_t^c}{(1+d)^t},
\end{equation}
where $d$ is the discount rate.

\subsection{Data \label{sec:data_generic}}

Our modelling approach is based on the following categories of data.
\begin{description}
	\item[Customer information:] general customer data including demographic data.
	\item[Product ownership:] information about the nature of a customer's product holding and information about product activity.
	\item[Customer activity and engagement:] features that track customer interaction. 
\end{description}
Note that only product ownership data is a strict requirement for our modelling approach, as they are required to segment customers as part of the model; additional data types mentioned above may be used to improve the overall performance of the model.
Product ownership and activity data in particular has been shown to be useful in predicting future purchases, see for example the RFM model developed by \citet{fader2005}.
We give more specific examples of data types used in application to retail banking in \Cref{sec:application}.

\subsection{Modelling approach \label{sec:modelling_approach}}
	
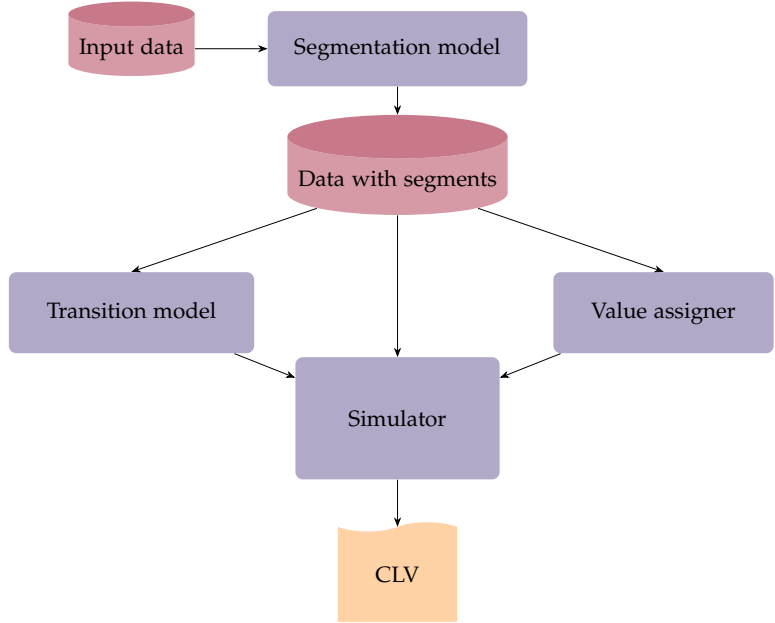
\begin{figure}
\begin{center}
\scalebox{0.7}{
\begin{tikzpicture}[shape aspect=0.2]
	\node[shape=cylinder, cylinder uses custom fill, shape border rotate=90, inner sep=0.2cm, cylinder end fill=myred, cylinder body fill=myred!75] (data) at (-5, 2){Input data};

	\node[shape=rectangle, rounded corners, inner sep=0.5cm, fill=mypurple] (seg_model) at (0, 2){Segmentation model};

	\node[shape=cylinder, cylinder uses custom fill, shape border rotate=90, inner sep=0.2cm, cylinder end fill=myred, cylinder body fill=myred!75] (seg) at (0, -0.5){Data with segments};

	\node[shape=rectangle, rounded corners, fill=mypurple, inner sep =0.5cm] (trans) at (-5, -3) {\begin{tabular}{c} Transition model\end{tabular}};

	\node[shape=rectangle, rounded corners, fill=mypurple, inner sep=0.5cm] (va) at (5, -3) {\begin{tabular}{c} Value assigner \end{tabular}};

	\node[shape=rectangle, rounded corners, fill=mypurple, inner sep=1cm] (simulation_1) at (0, -5) {Simulator};

	\node[shape=tape, tape bend bottom=none, fill=orange!35, inner sep=0.5cm] (clv_1) at (0, -8) {\begin{tabular}{c}  CLV \end{tabular}};

	\path[-Stealth] (data) edge (seg_model);
	\path[-Stealth] (seg_model) edge (seg.north);
	\path[-Stealth] (seg) edge (trans.north);
	\path[-Stealth] (seg) edge (simulation_1);
	\path[-Stealth] (seg) edge (va.north);
	\path[-Stealth] (va) edge (simulation_1);
	\path[-Stealth] (trans) edge (simulation_1);

	\path[-Stealth]  (simulation_1) edge (clv_1);

\end{tikzpicture}}
\caption{\label{fig:model_high_level} High-level overview of the CLV model structure. 
See \Cref{fig:model_training} for a detailed view of the structure of the CLV model.}
\end{center}
\end{figure}

For a given customer $c$ and time horizon $T$, our CLV model estimates the value of $CLV_T^c$, as defined in \cref{eq:clv} above.
The modelling framework extends that of \cite{Haenlein2007} by introducing machine learning modelling components in order to improve on predictive performance and allow predictions of CLV for longer time horizons. 
It is formed of four main steps, and a high-level view of how these steps fit together can be seen in \Cref{fig:model_high_level}. 
The modelling approach allows us to not only estimate the value of CLV but also anticipate specific customer needs related to product uptake, such as the likelihood of purchasing a home and requiring a mortgage within the next year. 
The four steps are described in the next sections as follows: 
segmentation model (\Cref{sec:seg}),
transition model (\Cref{sec:trans}),
value assigner (\Cref{sec:value}),
simulator (\Cref{sec:sim}).

\subsubsection{Segmentation model\label{sec:seg}}

Similar to \cite{Haenlein2007}, we use a decision-tree-based regression model as the first step, using the initial customer value $CV_0$ as the target variable. 
The leaves of this tree-based model segments customers into $S$ segments $\{1, 2, \dots, S\}$.
The leaf in which a given customer $c$ belongs defines their starting segment $s_0$.
This defined set of segments is kept constant for future time-steps, however year-on-year movement between individual segments is expected and modelled using a separate transition model described in \Cref{sec:trans}.
In order to develop a model that is product centred, and from which we can extract customer product needs, we require the use of product ownership data as input features to the model. 
We additionally introduce forced splits to the top of the tree to ensure splits on certain product holdings are used.
These forced splits facilitate a more interpretable model through segments that are clearly defined in terms of specific product holdings.
The nature of these forced splits is dependent on the use case and desired outcomes of the model, and we give details of how these can be chosen in \Cref{sec:application}.

\subsubsection{Transition model\label{sec:trans}}

In this step, we model the probability that a customer will move from a segment in one time period to each segment in the following time period.
Unlike \cite{Haenlein2007}, who use a first-order Markov decision process to give segment-level transition probabilities, we instead train a supervised machine learning model to give customer-level transition probabilities.
A feature vector containing customer $c$'s data (comprising the data described in \Cref{sec:data_generic}) at time-step $t - 1$ is denoted by $\mathbf{x}_{t - 1}^c$. The trained transition model takes as input the previous segment $s_{t - 1}$ and data $\mathbf{x}_{t - 1}^c$ and outputs probabilities $p(s_t = s\mid s_{t-1}, \mathbf{x}_{t - 1}^c)$ for $s\in \{1, \dots, S\}$ of moving from $s_{t - 1}$ to segment $s$ at time-step $t$. 
When predicting transitions from the starting period to the first period (i.e., $t=1$), there is a richer feature set available for $\mathbf{x}_0^c$, as we have data available for current product activity for example.
On the other hand, the feature vector $\mathbf{x}_{t - 1}^c$ for $t > 1$ (i.e., for transitions further into the future) may have smaller dimension, since model features are constrained to static features and those which can naturally be forward progressed such as customer age. 
As a consequence, two separate transition models are developed, a \emph{full} transition model for the first period using all available data and a \emph{simple} transition model for any period beyond the first where only a limited set of data attributes are known.

\subsubsection{Value assigner\label{sec:value}}

In this step, we predict a customer's value $v(s, \mathbf{x}_{t-1}^c)$ given the customer's segment $s_t = s$ at time-step $t$ and their available features $\mathbf{x}_{t-1}^c$. 
In contrast to \cite{Haenlein2007}, who assign value on a segment-level by taking the mean of the segment value, we train a supervised machine learning model to provide customer-level predictions of value for all segments $s$ that a customer may move to in the following time period. 
These customer-level predictions provide a more personalised view of each individual customer.
As with the transition model, the dimension of $\mathbf{x}_{t - 1}^c$ for $t > 1$ may be smaller than that of $\mathbf{x}_0^c$, being constrained to static features and features that can naturally be forward progressed. So, we have two value assigner models -- a \emph{full} value assigner for the first period and a \emph{simple} value assigner for periods beyond the first.

\subsubsection{Simulator\label{sec:sim}}

The three components of the model described above allow us to predict the first-period $CLV_1^c$ for customer $c$ as follows. 
We predict the first-period transition probabilities $p(s_1 = s\mid s_0, \mathbf{x}_0^c)$ for $s\in\{1, \dots, S\}$ using the trained full transition model as well as the customer's data $\mathbf{x}_0^c$ and starting segment $s_0$ as estimated by the segmentation model. We also predict the first-period customer values $v(s, \mathbf{x}_0^c)$ for all segments $s$ using the full value assigner.
The first-period prediction of both customer value and customer lifetime value is given as the expected value over the transition probability state:
\begin{equation}\label{eq:clvpred}
	\widetilde{CV}_1^c = \widetilde{CLV}_1^c = \sum_{s=1}^Sp(s_1 = s\mid s_0, \mathbf{x}_0^c)v(s, \mathbf{x}_0^c).
\end{equation}
To extend this to predict $CLV_T^c$ for longer time horizons, the final stage of the model runs a multi-period simulation. 
For example, the probability of customer $c$ moving to segment $s$ in the second period can be estimated as:
\begin{equation}\label{eq:trans2}
	p(s_2 = s\mid s_0, \mathbf{x}_0^c) = \sum_{r=1}^S p(s_2 = s\mid s_1 = r, \mathbf{x}_1^c)p(s_1 = r\mid s_0, \mathbf{x}_0^c).
\end{equation}
That is, the probability of a customer moving to segment $s$ in the second period can be estimated by summing the probabilities of moving to all the various segments in the first period, and then back to segment $s$ in the second period. 
The simple transition model is used to predict $p(s_2 = s\mid s_1 = r, \mathbf{x}_1^c)$ using $r$ and $\mathbf{x}_1^c$ as the input segment and customer data respectively.
We also use the simple value assigner to predict the second-period predicted customer values, $v(s, \mathbf{x}_1^c)$.
Prediction of $CV_2^c$ is given by:
\begin{equation}\label{eq:clvpred2}
	\widetilde{CV}_2^c = \sum_{s=1}^S p(s_2 = s\mid s_0, \mathbf{x}_0^c)v(s, \mathbf{x}_1^c).
\end{equation}
and $\widetilde{CLV}_2^c = \widetilde{CV}_1^c + \widetilde{CV}_2^c$.
This procedure may be repeated an arbitrary number of times to obtain $\widetilde{CLV}_T^c$. 

\section{Application to the retail banking industry\label{sec:application}}

In this section we provide results from a concrete application of the CLV modelling framework described in \Cref{sec:model} for customers at a large retail bank in the UK. 
In this application, we consider six core products -- current accounts, savings accounts, loans, credit cards, mortgages, and investments -- through which CLV is defined (\Cref{sec:clv_def}). 
In this case, the product cost $C_t^{c, p}$ for customer $c$ and product $p$ in year $t$, \Cref{eq:ev}, is approximated\footnote{Additional variable and fixed servicing costs were excluded, but may be included as per the user requirements.} as 
\begin{equation}\label{eq:cost}
	C_t^{c, p} = EL_t^{c, p} + CC_t^{c, p} + CR_t^{c, p},
\end{equation}
where $EL_t^{c, p}$ denotes the \emph{expected loss} as estimated by the sum of the possible losses weighted by the probability of loss occurrence, $CC_t^{c, p}$ denotes the \emph{cost of capital} as the estimated minimum return required to gain a profit, and $CR_t^{c, p}$ denotes the \emph{collections \& recoveries} given as any owed debt and expected debt recovery costs. 

Out of all customers at the bank, the \emph{Affluent} base is defined as those who satisfy the qualifying criteria for the bank's Premier banking program, and the \emph{Premier} base as the subset of \emph{Affluent} customers who are actually signed up to the program. 
By \emph{Retail} customers, we mean all customers of the bank's retail banking services who are not in the Affluent base. 
Finally, there is a separate database, \emph{Private}, for a private subsidiary bank which contains high-net-worth private banking clients.
This gives a set of three customer bases of increasing expected wealth and value: Retail, Affluent, and Private. 
Whilst CLV models have been implemented for all of these customer bases, in this section we focus on a CLV model developed specifically for Affluent customers. 
For confidentiality, monetary values have been normalised and are reported in \emph{Currency Units (CU)}.

\subsection{Business requirements and constraints}

Two key business requirements and constraints define the nature of the problem and influence specific design choices in model development.
\begin{enumerate}
	\item \textbf{Short data history.} Only three years of data history were available for this application. This means validation is only possible for one- and two-year CLV predictions. Whilst longer time horizons for predicting CLV were preferred, a decision to predict CLV for 5 years was made to balance longer prediction horizons with predictive performance.
	\item \textbf{Marketing campaign usage.} CLV model outputs will be used to support customer contact campaigns that are based around customer lifetime journeys and product uptake.
        Therefore, a degree of interpretability in the initial value-based segmentation of the customer base was required. 
	For example, for a proposed marketing campaign around the uptake of investment products it was necessary to be able to separate segments of customers that hold investment products already from segments of customers that do not hold an investment product.
	The likelihood of a customer to take up an investment product could then be estimated by their transition probabilities to segments corresponding to investment product holders.
\end{enumerate}

\subsection{Dataset\label{sec:data_training}}

The entire banking dataset used for this application corresponds to approximately 21 million retail customer accounts
between 2018 and 2021. 
Data from more than 30 different banking systems were aggregated and combined
into a single dataset that was used as input to the modelling and validation stages of this application. 
These systems captured data types across the three categories described in \Cref{sec:data_generic} aggregated at a monthly level. 
Some specific examples are as follows.
\begin{description}
	\item[Customer information:] personal data such as tenure with the bank, demographic data such as mean house price in the area, and customer risk data.  
	\item[Product ownership:] product holdings, balances, transactions and revenue for each of the six core products\footnote{To avoid double counting joint account data was given by assigning 50\% to each party.}. 
	\item[Customer activity and engagement:] tracking information such as pages visited on the bank's mobile app, customer marketing permissions and preferences, and customer complaints.
\end{description}
Features were annualised to match the chosen annual time period, which smooths out seasonal effects and reduces the size of the data.
The resultant input dataset provided high-dimensional data on each customer and contained around 980 million rows, reduced from over 400 billion rows of raw data originally.

\subsubsection{Exploratory analysis\label{sec:eda}}

\begin{figure}[H]
	\centering
	\includegraphics[trim=200 200 200 200, clip, width=0.9\textwidth]{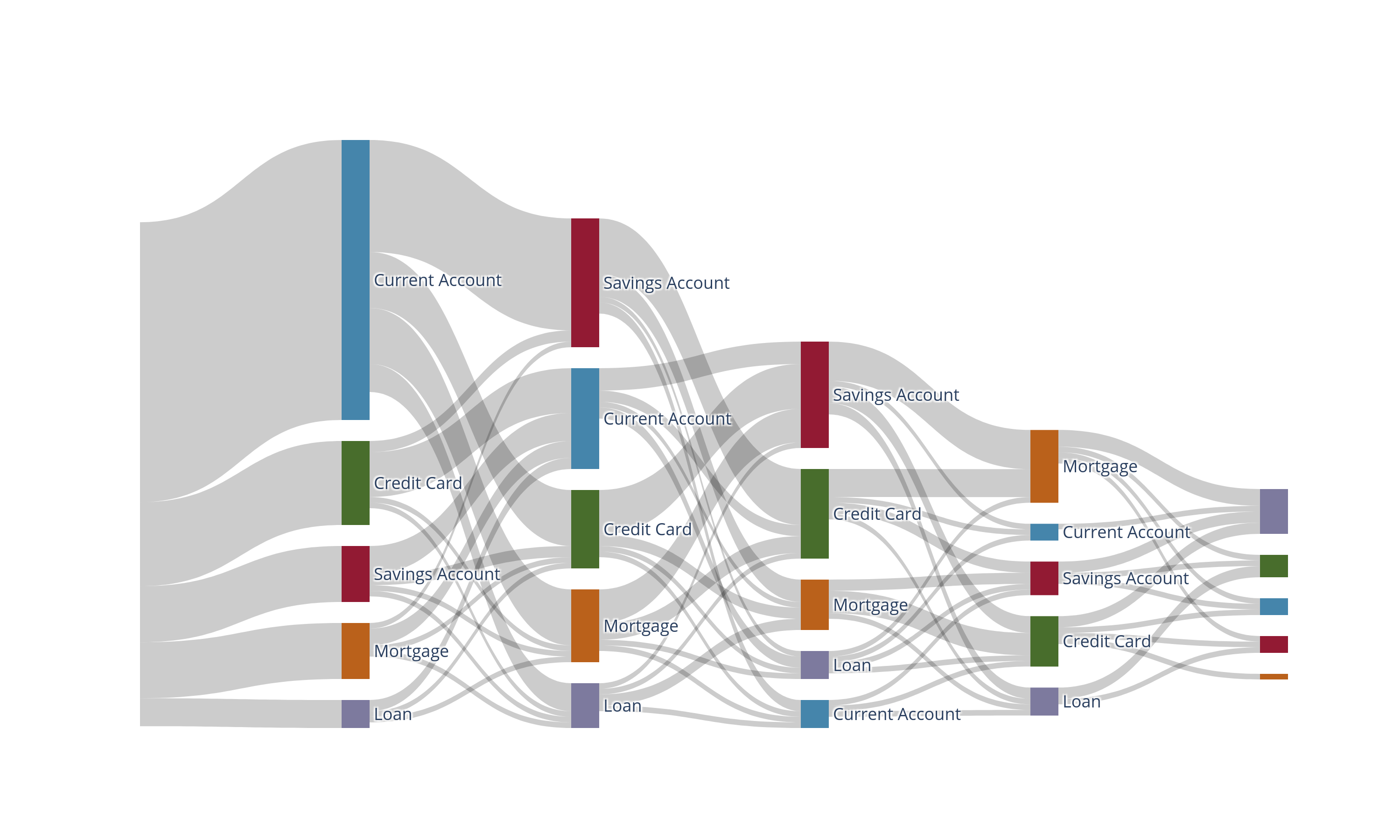}
	\caption{\label{fig:sankey_prod} Representative example of the order in which Premier customers acquired products. This is defined by the product acquisition date, so that the first column shows the first product acquired for all Premier customers. Subsequent columns show which product each customer went on to acquire next, if any, and so on.}
\end{figure}

In this section, we explore analyses of customer value in order to provide context and justification for design choices in the model that are detailed in \Cref{sec:model_results}.

\begin{figure}[h]
	\centering
	\includegraphics[trim=6 6 6 6, clip, width=0.8\textwidth]{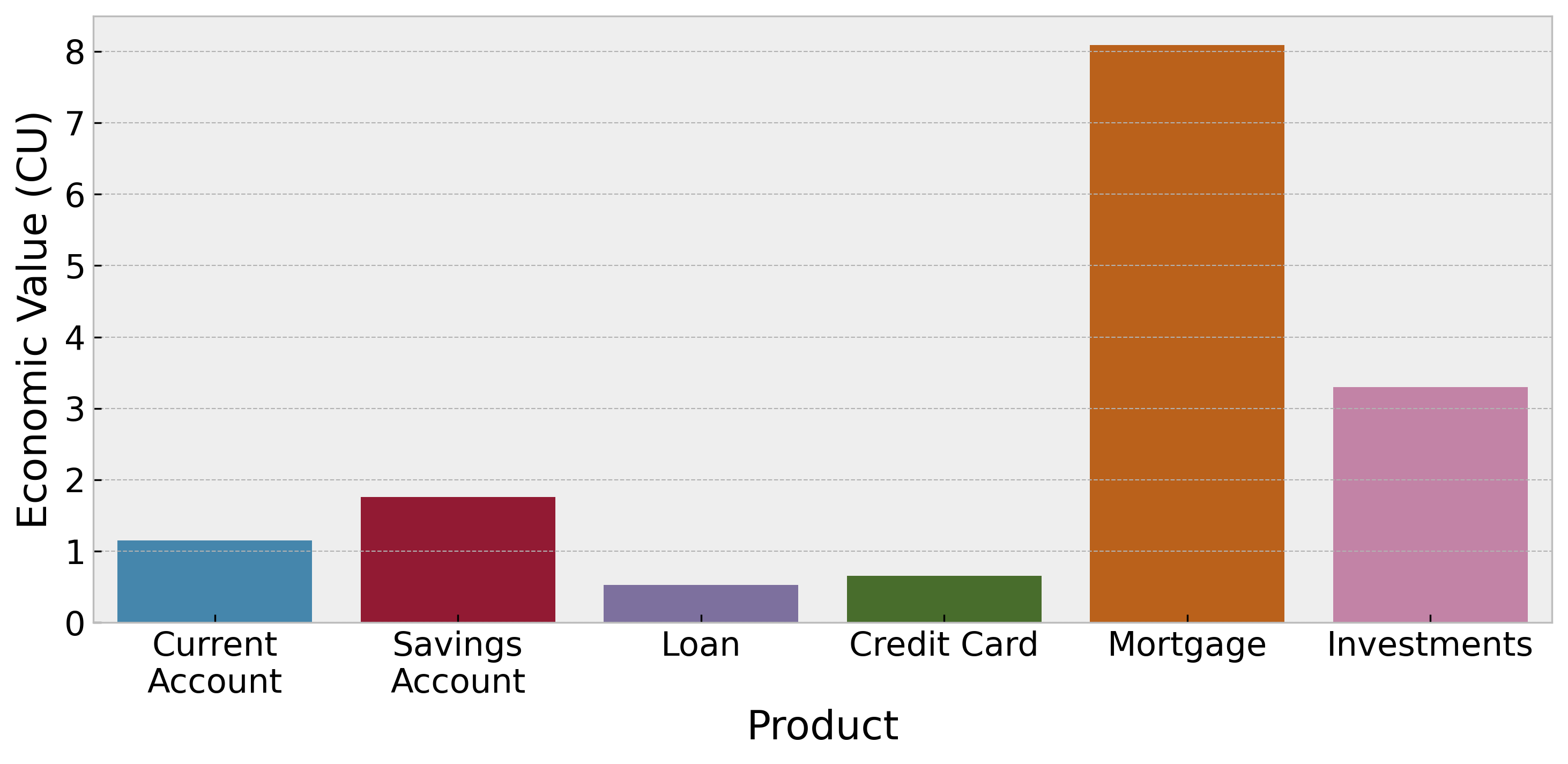}
	\caption{\label{fig:prod_value} Average value per product in currency units for Premier customers.}
\end{figure}

\Cref{fig:sankey_prod} is a representative illustration of the order in which Premier customers obtained products across their time with the bank. 
This indicates that modelling CLV in terms of product holdings is appropriate in order to capture changes in value over long and complex customer journeys. 
We can see that current accounts tend to be an early product, with savings, loans, mortgages and credit cards rising in prominence in later stages.
However not all products contribute equally to customer value as shown in the bar chart in \Cref{fig:prod_value}. 
A key issue we face here is the large disparity in value between the highest value product (mortgages) and the other products.
The result of this, as we discuss in the next section, is that if we let the segmentation tree develop greedily, mortgage holdings and balances tend to dominate all splits. 
This leads to segments that are not clearly defined on product holdings for products other than mortgages, which motivates the need for forced splits.

\begin{figure}[h]
	\centering
	\includegraphics[trim=10 10 50 10, clip, width=0.85\textwidth]{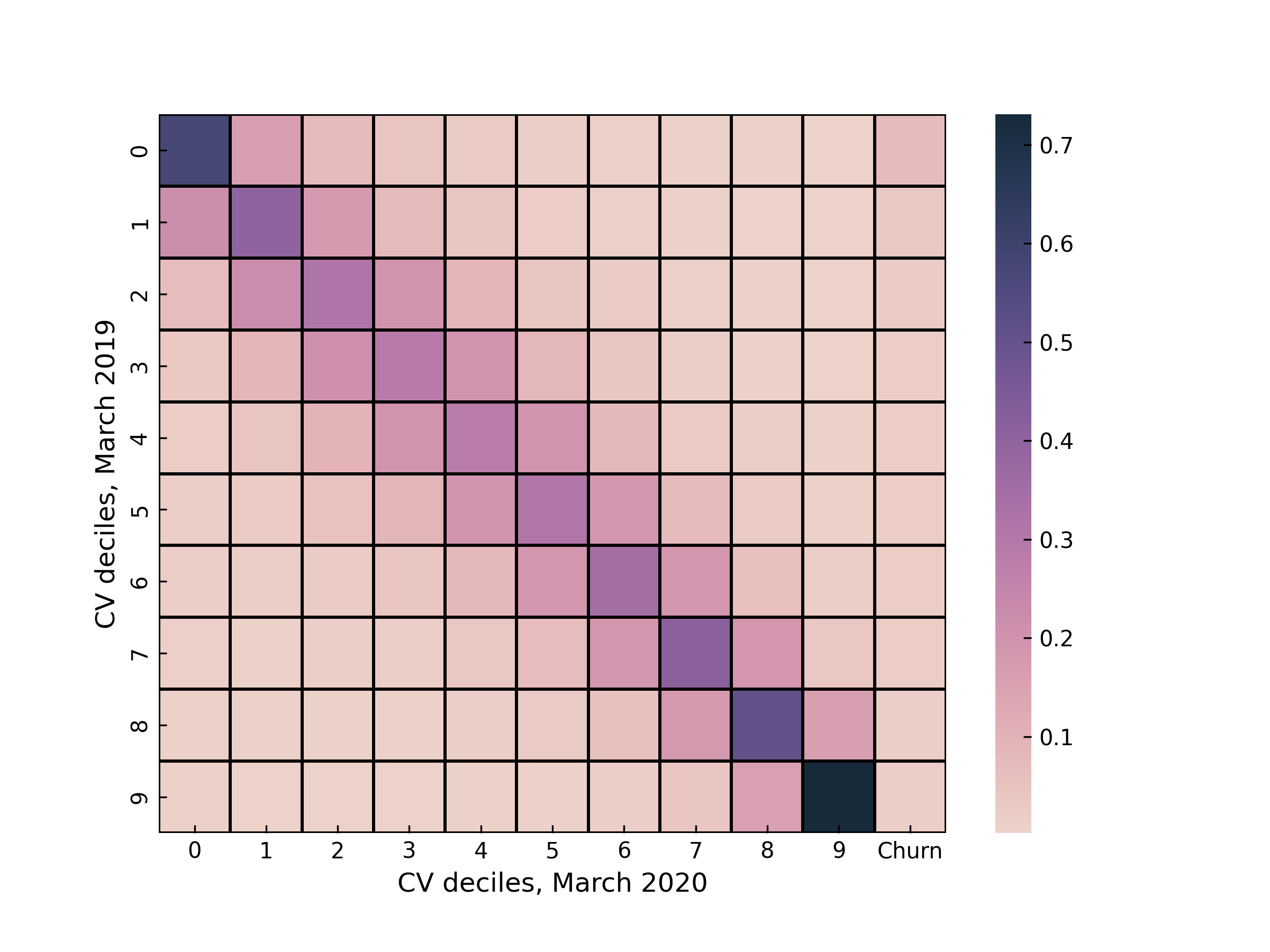}
	\caption{\label{fig:deciles} Observed CV decile transition heatmap for Affluent customers between March 2019 and March 2020. Deciles are ordered so that Decile 0 corresponds to the lowest-value 10\% of customers and Decile 9 corresponds to the highest-value 10\% of customers.}
\end{figure}

It is important that our definition of customer value captures expected and known dynamics in the customer base.
We observed a large amount of movement of customers across CV deciles between 2019 and 2020, with around 55\% of customers moving decile in this time period, demonstrating that on the time span of a single year, our CV estimates reflect the expected dynamics in the customer base. A more granular view on the transition proportions across CV deciles can be seen in \Cref{fig:deciles} from which we can see that a large proportion of transitions are driven by small changes and those in the middle CV deciles are more likely to change deciles than those in the lower or higher deciles.

\begin{figure}[H]
	\centering
\includegraphics[trim=6 6 6 6, clip, width=0.85\textwidth]{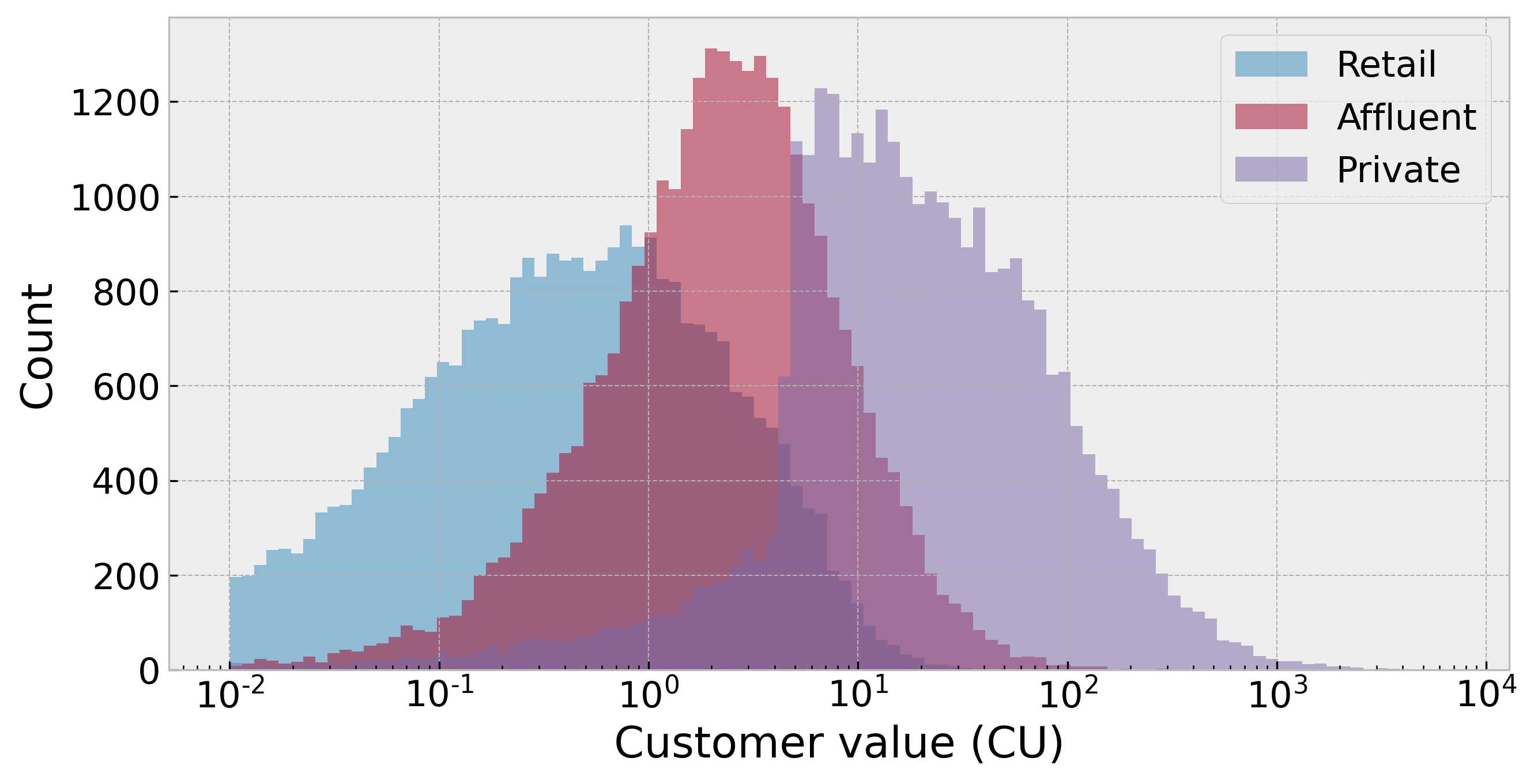}
	\caption{\label{fig:cv} Distribution of customer value in currency units on a $\log_{10}$ scale for customers in the dataset for the year 2020,
	split by customer base. Due to the log scale, this distribution plot demonstrates substantial shifts in customer value estimates across the bases.}
\end{figure}

\Cref{fig:cv} illustrates how the distribution of customer value in 2020 varies with wealth. 
In particular, we see a noticeable and expected shift in customer value across the three different customer bases of increasing affluence -- Retail, Affluent and Private.

\begin{figure}[h]
	\centering
\includegraphics[trim=0 0 25 25, clip, width=\textwidth]{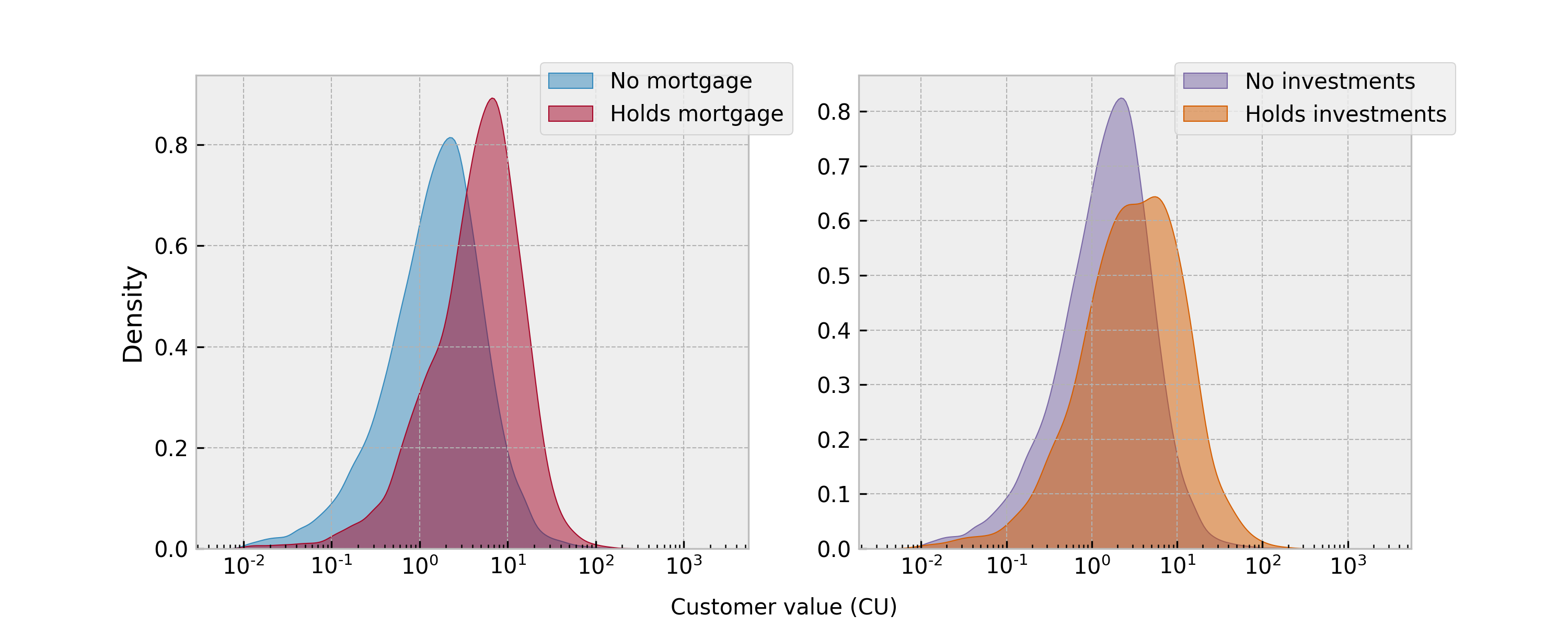}
	\caption{\label{fig:cv_prod_dist} Distribution of customer value in currency units on a $\log_{10}$ scale for Affluent customers in the dataset for the year 2020 according to whether customers held mortgages/investments or not. 
	Customer values for mortgage and investment holdings are higher than their respective no holdings counterpart.}
\end{figure}

 In \Cref{fig:cv_prod_dist} the distribution of customer value in the Affluent base are shown according to mortgage and investment holdings which, as we saw in \Cref{fig:prod_value}, are the two highest-value products in the Affluent base. We see that customer value distributions separate according to such product holdings, making these features suitable for use in forced splits at the top of the segmentation tree.

\subsection{Model training and validation}
\label{sec:model_results}
In this section, we describe the specific model trained for our application along with the results that were observed in testing. 
Prior to model training, privacy impact and model fairness assessments were undertaken in order to identify and minimise data privacy risks and bias in the deployment of the CLV model.

\subsubsection{Model training}

Before training, we first filtered to Affluent customers and annualised the data set. Additional filters were applied on the dataset before input into the segmentation model. 
For testing, a random selection of 25\% of customers were held out in a test set to facilitate in-time and out-of-time validation. 
The model was subsequently trained on approximately 2,000,000 rows corresponding to around 800,000 customers, with around 700,000 rows corresponding to 275,000 customers being held out in the test set.
We now detail the implementation of the four modelling steps described in \Cref{sec:modelling_approach}.

\begin{enumerate}
	\item\textbf{Segmentation model.} 
        The segmentation tree was obtained by training a variant of a LightGBM Regressor\footnote{\href{https://www.microsoft.com/en-us/research/wp-content/uploads/2017/11/lightgbm.pdf}{https://www.microsoft.com/en-us/research/wp-content/uploads/2017/11/lightgbm.pdf}} with a single estimator on a single-year subset of the training data, with current customer value as the target variable. 
        Product holdings and transactions data, along with demographic data, formed the input features, however age features were specifically excluded since they correlate with CV and we sought primarily to segment based on product features. 
	Whilst we want to obtain segments that distinguish different value brackets as well as possible, they should also be interpretable to some degree so that they can be used to support specific marketing campaigns.
	In experiments we observed that letting the tree train without forced splits led to all splits being based on mortgage product data, which would make it difficult to extract product holdings from the resultant segments for other products, such as investments.
	Therefore we used forced splits based on product holdings at the top of the tree so that segments with or without products are clearly defined.
	To choose these forced splits we use higher-value products that are indicative of customer value, obtained from analyses such as \Cref{fig:prod_value} and \Cref{fig:cv_prod_dist}, as well as any specific needs from proposed marketing campaigns. 

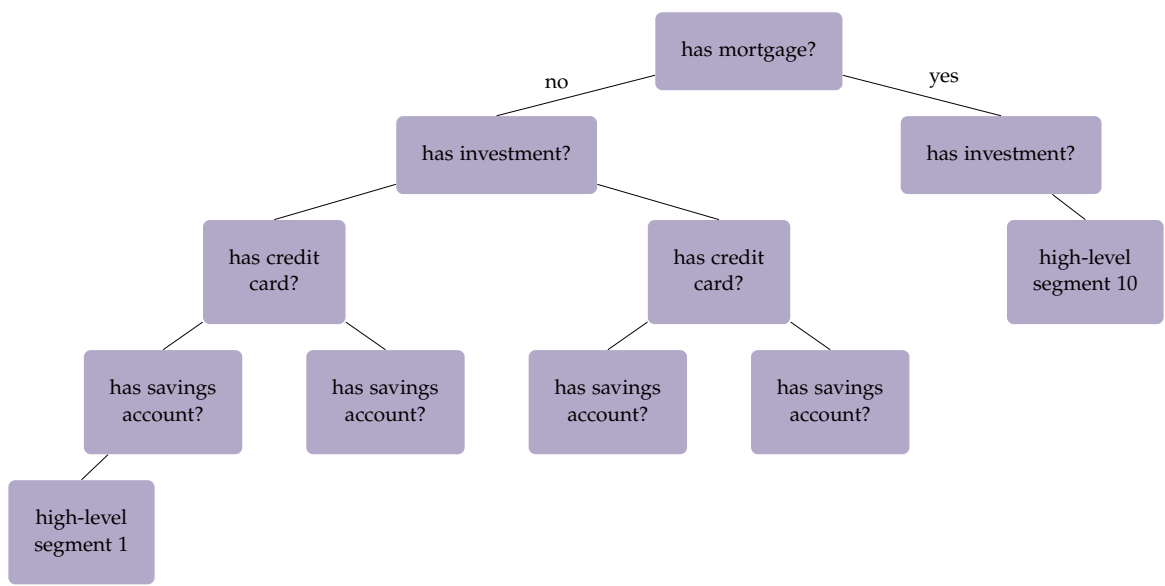
\begin{figure}[h]
\begin{center}
\scalebox{0.65}{
\begin{forest}
for tree={
	rounded corners,
	inner sep=15pt,
	fill=mypurple,
	base=bottom,
	child anchor=north,
	align=center,
	s sep+=1cm,
	},
[has mortgage?
	[has investment?, edge label={node[midway, above left]{no}}
		[has credit \\ card?
			[has savings \\ account?
				[high-level \\ segment 1]
				[, phantom]
			]
			[has savings \\ account?]
		]
		[has credit \\ card?
			[has savings \\ account?]
			[has savings \\ account?]
		]
	]
	[, phantom]
	[, phantom]
	[has investment?, edge label={node[midway, above right]{yes}}
		[, phantom]
		[high-level \\ segment 10]
	]
]
\end{forest}
}
\caption{\label{fig:forced_splits} Forced splits used in the segmentation tree. These forced splits lead to ten high-level segments, two of which are labelled. High-level segment 1 corresponds to all customers that hold no mortgage, no investment, no credit card and no savings account. High-level segment 10 corresponds to all customers that hold a mortgage and investments.}
\end{center}
\end{figure}

\begin{figure}[h]
	\centering
\includegraphics[trim=6 6 6 40, clip, width=0.85\textwidth]{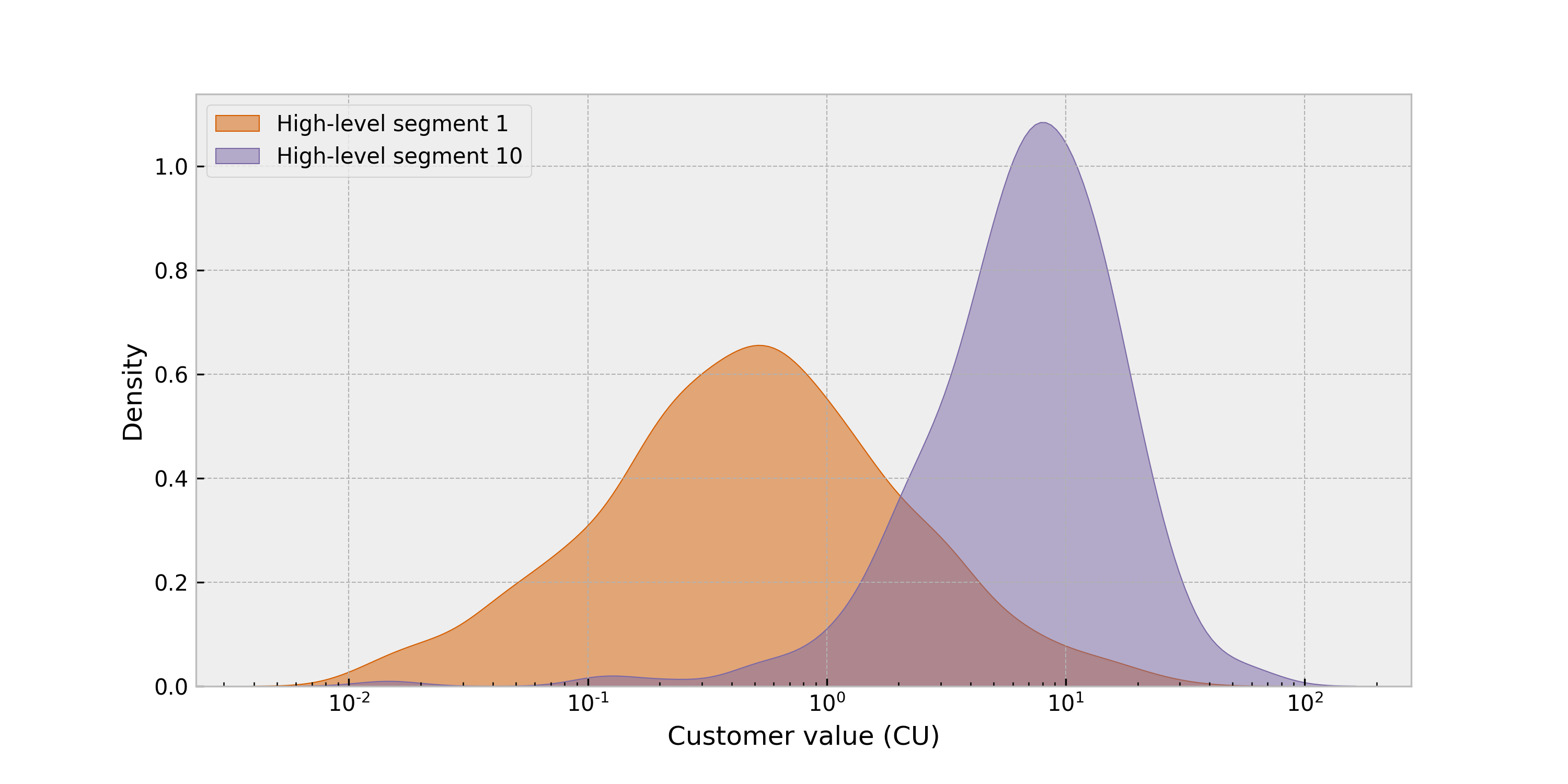}
	\caption{\label{fig:forced_split_dist} Distribution of customer value in currency units on a $\log_{10}$ scale for customers in high-level segment 1 vs. those in high-level segment 10, which correspond respectively to the lowest- and highest-value segments from \Cref{fig:forced_splits}.}
\end{figure}

All of the forced splits used can be seen in \Cref{fig:forced_splits}. These forced splits define ten high-level segments of customers based on the various product holdings. In \Cref{fig:forced_split_dist}, we can see the difference in value distribution of the lowest-value segment vs. the highest-value segment from \Cref{fig:forced_splits}, demonstrating that the choice of forced splits is capable of separating value brackets.
Once the forced splits are defined, the remaining segmentation of customers is achieved by allowing the model to train greedily.
	Note that the right subtree in \Cref{fig:forced_splits} of customers who hold a mortgage, includes fewer forced splits than the left subtree consisting of customers who do not hold a mortgage.
	The high value of mortgage products, along with the fact that a majority of mortgage customers hold no other product, meant that this structure improved model performance compared with having the same forced splits on the right as on the left.
	Moreover, the relative lack of forced splits on the right subtree maintained the interpretability necessary for planned customer contact campaigns.
	Whilst higher numbers of segments can provide a more granular segmentation of the customer base in terms of current CLV, transitions amongst them become harder to predict. 
	To balance this, the number of segments was empirically selected at 50. 
	Other hyperparameters were tuned using a randomised search.  
	The eventual trained tree with 50 leaves, corresponding to the 50 segments, included splits based on 16 product-based features and a single demographic feature. 
	The 50th segment corresponded to customer churn.
	\item\textbf{Transition model.} 
        We trained both a \emph{full} transition model for first-year predictions and a \emph{simple} transition model for predictions beyond first-year. 
        Both transition models were multi-class LightGBM Classifier models, with the segment that a customer moves to in the following year as the target variable. 
        The training data included the segments for each customer and each year, as determined by the output of the segmentation model. 
        For the full transition model, we selected the 50 most important features amongst all relevant features; importance was determined by training the model on a small subsample of the data and extracting the importances from the underlying LightGBM models. 
        The simple transition model used a fixed set of 30 features only, which include static features and those features for which it was possible to forward simulate, such as age and product tenure. 
        The output of the transition model is a vector of 50 probabilities with the $i$th entry being the probability that a customer will move into segment $i$ in the following year. 
	\item\textbf{Value assigner.} 
        Like the transition model, we trained a full and a simple value assigner model, which were both LightGBM Regressor models using prediction-year customer value as the target variable. 
        The input data included the segments for each customer and each year as determined by the segmentation model. 
        The full value assigner included all product holdings and transactions features, demographic data, as well as the 1-year lagged target variable. 
        The simple value assigner had only a fixed subset of 25 features whose values were easy to simulate year-on-year. 
	\item\textbf{Multi-year simulation.} 
        The multi-year simulator used the trained transition models and value assigners to generate CLV predictions per customer per year, for all five years in prediction horizon, as described in \Cref{sec:sim}. 
        In our application this was a five-year prediction horizon, however as implemented the simulator is theoretically able to predict for an arbitrary time horizon. 
        The simulator used the full pre-trained transition and value assigner models for the prediction of first-year CLV, and subsequently used the simple pre-trained models for the remaining four CLV predictions. 
        Between each year, a specified set of features (i.e., those that are used as input for the simple models) were progressed in the natural manner (for example, adding 1 or 12 to yearly or monthly features respectively) in order to provide input data to the subsequent simple transition and value assigner models.
\end{enumerate}

\begin{figure}
\begin{center}
\scalebox{0.8}{
\begin{tikzpicture}[shape aspect=0.2]
	\node[shape=cylinder, cylinder uses custom fill, shape border rotate=90, inner sep=0.2cm, cylinder end fill=myred, cylinder body fill=myred!75] (master) at (0, 11){Master table};
	\node[shape=rectangle, rounded corners, inner sep=0.5cm, fill=mypurple] (model_input) at (0, 9.2){Preprocessing};
	\node[shape=cylinder, cylinder uses custom fill, shape border rotate=90, inner sep=0.2cm, cylinder end fill=myred, cylinder body fill=myred!75] (data) at (0, 7){Input data};
	\node[shape=rectangle, rounded corners, inner sep=0.5cm, fill=mypurple] (seg_model) at (0, 5){Segmentation model};
	\node[shape=cylinder, cylinder uses custom fill, shape border rotate=90, inner sep=0.2cm, cylinder end fill=myred, cylinder body fill=myred!75] (seg) at (0, 2){\begin{tabular}{c} Data with \\ initial segments\end{tabular}};
	\node[shape=rectangle, rounded corners, fill=mypurple] (full_trans) at (-5, 0) {\begin{tabular}{c} Full \\ transition model\end{tabular}};
	\node[shape=rectangle, rounded corners, fill=mypurple] (full_va) [below=3pt of full_trans] {\begin{tabular}{c} Full \\ value assigner \end{tabular}};
	\begin{scope}[on background layer]
	\node[shape=rectangle, rounded corners, fill=mypurple!35, fit=(full_trans)(full_va)] (full_models){};
	\end{scope}

	\node[shape=rectangle, rounded corners, fill=mypurple] (simple_trans) at (5, 0) {\begin{tabular}{c} Simple \\ transition model\end{tabular}};
	\node[shape=rectangle, rounded corners, fill=mypurple] (simple_va) [below=3pt of simple_trans] {\begin{tabular}{c} Simple \\ value assigner\end{tabular}};
	\begin{scope}[on background layer]
	\node[shape=rectangle, rounded corners, fill=mypurple!35, fit=(simple_trans)(simple_va)] (simple_models){};
	\end{scope}

	\node[shape=rectangle, rounded corners, rotate=270, fill=mypurple, inner sep=1cm] (simulation_1) at (-5, -6) {Simulator};
	\node[shape=cylinder, cylinder uses custom fill, shape border rotate=180, rotate=270, cylinder end fill=myred, cylinder body fill=myred!75] (update_1) at (-3, -6) {Progressed data};
	\node[shape=rectangle, rounded corners, rotate=270, fill=mypurple, inner sep=1cm] (simulation_2) at (-1, -6) {Simulator};
	\node[shape=cylinder, cylinder uses custom fill, shape border rotate=180, rotate=270, cylinder end fill=myred, cylinder body fill=myred!75] (update_2) at (1, -6) {Progressed data};
	\node[shape=rectangle, rounded corners, rotate=270, fill=mypurple, inner sep=1cm] (simulation_3) at (3, -6) {Simulator};
	\node[shape=cylinder, cylinder uses custom fill, shape border rotate=180, rotate=270, cylinder end fill=myred, cylinder body fill=myred!75] (update_3) at (5, -6) {Progressed data};
	\node[shape=rectangle, rounded corners, rotate=270, fill=mypurple, inner sep=1cm] (simulation_4) at (7, -6) {Simulator};
	\node[shape=cylinder, cylinder uses custom fill, shape border rotate=180, rotate=270, cylinder end fill=myred, cylinder body fill=myred!75] (update_4) at (9, -6) {Progressed data};
	\node[shape=rectangle, rounded corners, rotate=270, fill=mypurple, inner sep=1cm] (simulation_5) at (11, -6) {Simulator};

	\node[shape=tape, tape bend bottom=none, fill=orange!35] (clv_1) at (-5, -10) {\begin{tabular}{c} 1-year \\ CLV \end{tabular}};
	\node[shape=tape, tape bend bottom=none, fill=orange!35]  (clv_2) at (-1, -10) {\begin{tabular}{c} 2-year \\ CLV \end{tabular}};
	\node[shape=tape, tape bend bottom=none, fill=orange!35]  (clv_3) at (3, -10) {\begin{tabular}{c} 3-year \\ CLV \end{tabular}};
	\node[shape=tape, tape bend bottom=none, fill=orange!35]  (clv_4) at (7, -10) {\begin{tabular}{c} 4-year \\ CLV \end{tabular}};
	\node[shape=tape, tape bend bottom=none, fill=orange!35] (clv_5) at (11, -10) {\begin{tabular}{c} 5-year \\ CLV \end{tabular}};

	\path[-Stealth] (master) edge (model_input.north);
	\path[-Stealth] (model_input) edge (data.north);

	\path[-Stealth] (data) edge (seg_model.north);
	\path[-Stealth] (seg_model) edge (seg.north);

	\path[-Stealth] (seg) edge (full_models.north);
	\path[-Stealth] (seg) edge (simulation_1);
	\path[-Stealth] (seg) edge (simple_models.north);
	\path[-Stealth] (full_models) edge (simulation_1.west);
	\path[-Stealth] (simple_models) edge (simulation_2.west);
	\path[-Stealth] (simple_models) edge (simulation_3.west);
	\path[-Stealth] (simple_models) edge (simulation_4.west);
	\path[-Stealth] (simple_models) edge (simulation_5.west);
	
	\path[-Stealth] (simulation_1) edge (update_1);
	\path[-Stealth] (update_1) edge (simulation_2);
	\path[-Stealth] (simulation_2) edge (update_2);
	\path[-Stealth] (update_2) edge (simulation_3);
	\path[-Stealth] (simulation_3) edge (update_3);
	\path[-Stealth] (update_3) edge (simulation_4);
	\path[-Stealth] (simulation_4) edge (update_4);
	\path[-Stealth] (update_4) edge (simulation_5);

	\path[-Stealth]  (simulation_1) edge (clv_1);
	\path[-Stealth] (simulation_2) edge (clv_2);
	\path[-Stealth] (simulation_3) edge (clv_3);
	\path[-Stealth] (simulation_4) edge (clv_4);
	\path[-Stealth] (simulation_5) edge (clv_5);
\end{tikzpicture}}
\caption{\label{fig:model_training} CLV model training and generation of CLV prediction for time horizon of five years.}
\end{center}
\end{figure}
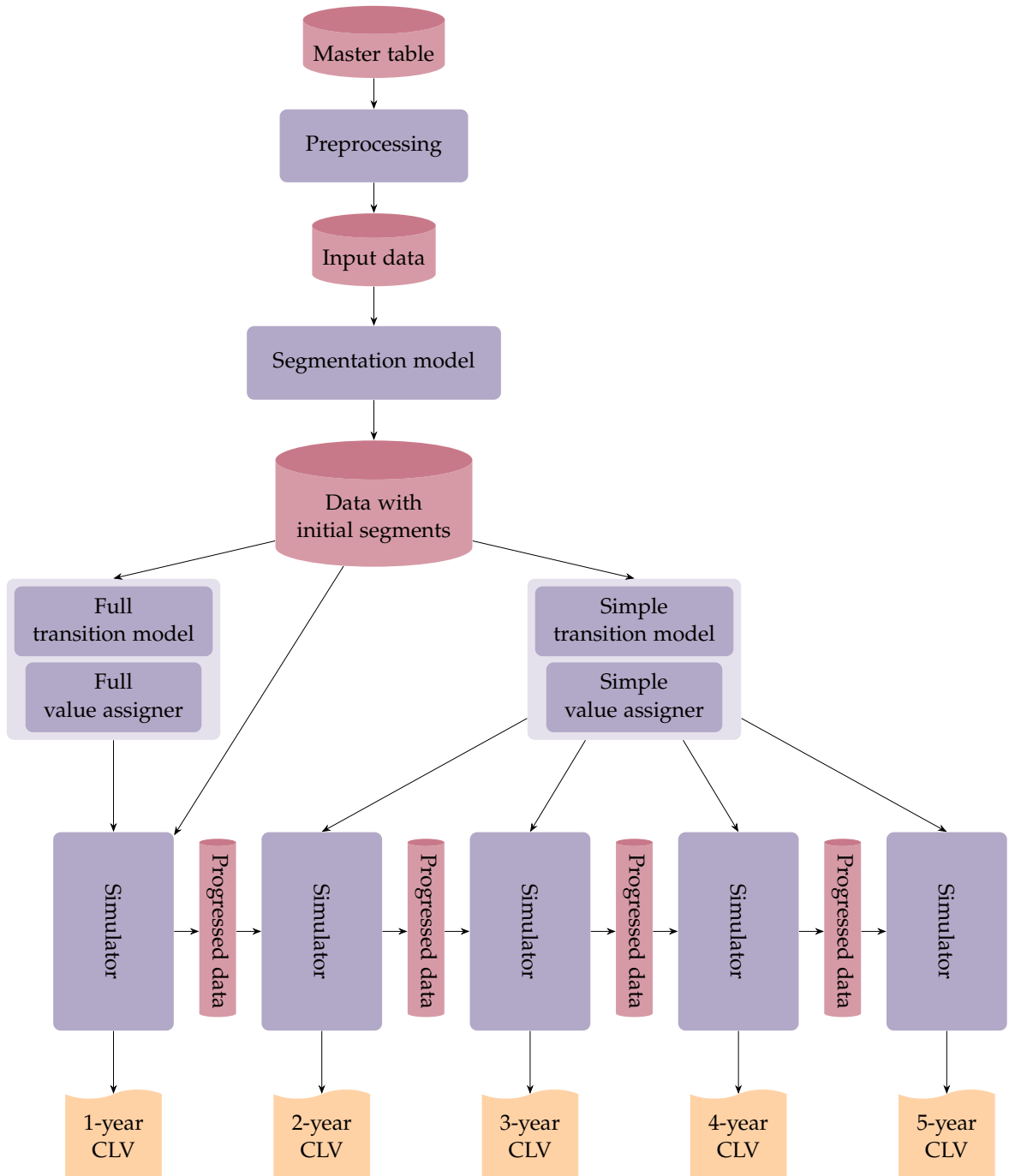

\paragraph{Model pipeline.} Model components were trained in a certain order due to the dependence of some models on the outputs of others. 
The segmentation model was trained first in order to generate customer segments, which could be attached to the original training data. 
Afterwards, these segments were used as inputs to train all four transition and value assigner models, which could be done in any order. 
Once these models were trained, the simulation step was performed last in order to generate final model outputs. 
Model training and prediction generation is summarised in \Cref{fig:model_training}.

\subsubsection{Model validation}
\label{sec:model_val}

Given the short data history, we were only able to validate the model's performance for one- and two-year CLV predictions. 
Specifically, the transition and value assigner models were trained on data spanning two years 2018--2020, and we validated the performances of the CLV model in predicting one-year CV for customers in the held-out test set in both 2019--2020 and 2020--2021 time periods, as well as two-year CV in the 2019--2021 time period, therefore providing both in-time and future out-of-time validation. 
Validation metrics were chosen to measure three core capabilities: \textbf{(1)} ability to predict the monetary CV value; \textbf{(2)} ability to rank customers by their predicted values; and \textbf{(3)} ability to predict specific customer needs. 
The latter two are inspired by marketing campaign requirements of the model, as the outputs of the model are used to rank customers and top deciles are chosen as targets to receive customer contact.
Further validation is provided through model usage in live campaigns, see \Cref{sec:inter}.
We detail specific metrics used to measure capabilities \textbf{(1)}--\textbf{(3)} below.

\begin{description}
\item{\textbf{(1) Median Absolute Error.}} To measure the models ability to predict monetary CV values we used the \emph{Median Absolute Error} (\emph{MedAE}) of the model's customer value predictions for a given year as compared with the observed values.

\item{\textbf{(2a) Separation.}} The \emph{Separation} metric is used to measure ranking capability of the model. This metric measures the observed CV uplift in the top $x$ percent of customers compared to the bottom $x$ percent, as ranked by their predicted CV. Specifically, we have 
	\begin{equation}\label{eq:sep}
		\text{\emph{Separation}}_{x} = \frac{\text{mean actual $CV_t$ of $T_{t, x}$}}{\text{mean actual $CV_t$ of $B_{t, x}$}},
	\end{equation}
	where $T_{t, x}$ is the top $x$\% of customers ranked by predicted $\widetilde{CV}_t$, and $B_{t, x}$ is the bottom x\% of customers ranked by predicted $\widetilde{CV}_t$. 
    This tells us how many times higher in value our top predicted $x$\% actually is, in comparison to the bottom predicted $x$\%, and therefore how good our model is at separating the higher-value customers from the lower-value customers.

\item{\textbf{(2b) Top-$x$ Precision.}} To measure ranking capability, we also considered the \emph{Top-x Precision} for year $t$, which is the share of top $x$\% of customers ranked by predicted $\widetilde{CV}_t$ that are in the top $x$\% of customers when ranked by actual $CV_t$.
In other words, it is the precision score of the binary classification task obtained by letting the ground truth label be 1 if the customer has $CV_t$ in the top $x$\% of actual CV and 0 otherwise, and the prediction label be 1 if the customer has predicted $\widetilde{CV}_t$ in the top $x$\% of predicted CV and 0 otherwise. 

\item{\textbf{(3a) Accuracy (50 Class).}}
            To measure the ability to predict customer needs, we measure how well our transition model is at predicting future segments. 
            The transition model outputs are converted to 50-class classification outputs by predicting a customer's next-year segment to be the segment of the highest transition probability. 
            We then used the accuracy score of this classification task to evaluate the transition model.
\item{\textbf{(3b) Accuracy (4 Class).}}
            Restricting Accuracy (50 Class) in (3a) to the first two levels of the tree measures how well the model is at specifically predicting mortgage and investment product uptake, which we call \emph{Accuracy (4 Class)} since there are 4 classes in this case.
            When analysing specific campaign effectiveness using the CLV model, there are additional metrics we used to measure the ability of the model to predict investment product uptake, which will be described in detail in \Cref{sec:inter}.
\end{description}

The performances of our model with respect to the above metrics for all validation time periods are compared to a Markov baseline model which is derived from \ \cite{Haenlein2007}.
This model used the same segmentation first step as our CLV model. 
The baseline transition model is given by a first-order Markov process, where transition probabilities are estimated via historical frequencies; that is, the probability of a customer moving from segment $s$ to segment $r$ is estimated by the relative frequency of customers in segment $s$ who moved to segment $r$ as given by the training data. 
The baseline value assigner is given by a mean value assigner, which predicts the value of a customer in segment $s$ as simply by the average of customer value in that segment. 
Finally, the baseline simulator generates predictions in the same manner that our CLV simulation model does, but with the Markov transition model and mean value assigner in place of the LightGBM transition and value assigner models respectively.
In \Cref{table:results_oneyear}, we see that our modelling approach achieves improvements over the Markov baseline model across the board.
In particular, we see a large improvement in Accuracy (50 Class), which further demonstrates the success of our approach of using machine learning models for the transition model and value assigner.

\begin{table}
\begin{center}
\caption{\label{table:results_oneyear} Evaluation of our CLV model compared to the Markov baseline in predicting one- and two-year in-time and out-of-time CLV values. Specific definitions for the metrics can be found in \Cref{sec:model_val}. Higher values are preferable for all metrics with the exception of MedAE. $x$ denotes the top percentage of customers chosen when ranked by predicted CV.}
\resizebox{\textwidth}{!}{
\begin{tabular}{|c|c|c|c|c|c|c|c|}
	\hline
	\multicolumn{2}{|c|}{\multirow{2}{*}{\textbf{Metric}}}   & \multicolumn{2}{|c|}{\begin{tabular}{c} 1-year in-time \\ (2019--2020)\end{tabular}} & \multicolumn{2}{|c|}{\begin{tabular}{c}2-year\\ (2019--2021)\end{tabular}} & \multicolumn{2}{|c|}{\begin{tabular}{c} 1-year out-of-time \\ (2020--2021)\end{tabular}} \\
	\cline{3-8}
	\multicolumn{2}{|c|}{} & CLV & Baseline & CLV & Baseline & CLV & Baseline \\
	\hline
	\multicolumn{2}{|c|}{\textbf{(1) MedAE}} & 1.05 & 1.54 & 1.55 & 1.90 & 0.86 & 1.52 \\
	\hline
	\multirow{3}{*}{$\textbf{(2a) Separation}_x$} & $\bm{x = 10}$ & 32 & 13 & 17 & 12 & 42 & 14 \\
	\cline{2-8}

	& $\bm{x=20}$ & 17 & 10 & 11 & 9 & 21 & 10 \\
	\cline{2-8}
	& $\bm{x=40}$ & 8 & 5 & 6 & 4 & 8 & 5\\
	\hline 
	\multirow{3}{*}{\textbf{(2b) Top-\emph{x} Precision}} & $\bm{x=10}$ & 0.67 & 0.54 & 0.55 & 0.53 & 0.71 & 0.53 \\
	\cline{2-8}
	& $\bm{x=20}$ & 0.71 & 0.61 & 0.63 & 0.60 & 0.74 & 0.65 \\
	\cline{2-8}
	& $\bm{x=40}$ & 0.78 & 0.65 & 0.68 & 0.63 & 0.80 & 0.68 \\
	\hline
	\multicolumn{2}{|c|}{\textbf{(3a) Accuracy (50 Class)}} & 0.79 & 0.57 & 0.54 & 0.44 & 0.77 & 0.55\\
	\hline
	\multicolumn{2}{|c|}{\textbf{(3b) Accuracy (4 Class)}} & 0.95 & 0.93 & 0.89 & 0.88 & 0.94 & 0.94 \\
	\hline
\end{tabular}
}
\end{center}
\end{table}

\subsubsection{Ethical considerations}

When developing a CLV model for an application via our proposed modelling methodology, care should be taken to address ethical issues. 
Since the model is trained on historical data, it may learn to recreate biases present within the data. 
This may be addressed in data engineering processes prior to input to the model, or through the addition of constraints on model training. 
Moreover, feature selection should avoid features that target protected groups, either directly or indirectly. 
For example, in our application we specifically excluded features based on gender and nationality, amongst others.
Finally, it is important to validate the bias and fairness of the model before using it in production.

\subsection{Customer contact campaigns\label{sec:inter}}

In this section we describe how the CLV model is being used to support customer contact campaigns around specific products. 
As an example, we consider a customer contact campaign based on investment products.

We can derive investment propensity models from our CLV model due to the use of forced splits in the segmentation model. 
Since mortgage holdings and investment product holdings form the first- and second-level splits, respectively, of the segmentation tree we obtain four subtree splits $S_{00}, S_{01}, S_{10}, S_{11}$ corresponding respectively to customers who hold: no mortgage and no investment products; no mortgage and investment product; mortgage and no investment products; mortgage and investment products. 
A customer with no investments belongs to either $S_{00}$ or $S_{10}$ and, using the pre-trained transition model, we can compute their investment propensity for the next year as the sum of transition probabilities over all segments in investment-holding subtrees $S_{01}$ and $S_{11}$.

We evaluated the test-set performance of the one-year investment propensity model using lift curves and lift values on the period June 2019 -- May 2020. 
That is, on June 2019 we generate the propensity that each customer will obtain an investment product by May 2020, and we compare this to the ground truth via the lift curve.  

\begin{figure}[h]
	\begin{center}
	\includegraphics[width=0.7\textwidth]{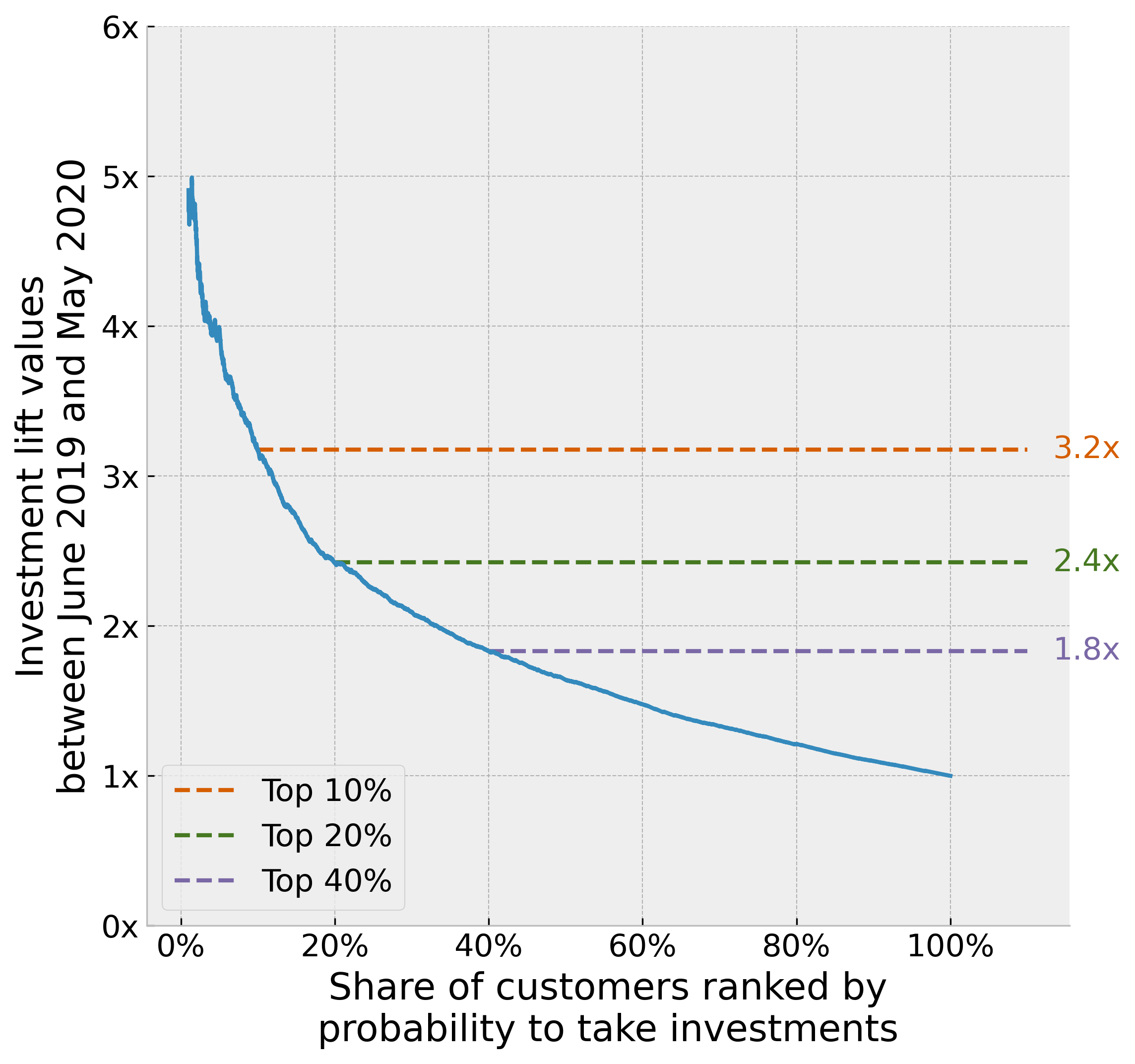}
	\end{center}
	\caption{\label{fig:lift} One-year lift curve for the investment propensity model applied to Premier customers.}
\end{figure}
 
The lift curve for the investment propensity model applied to the subset of Premier customers is illustrated in \Cref{fig:lift}.
The $x$-axis gives the top $x$\% of customers when ranked by predicted propensity and the $y$-axis plots the corresponding lift value, defined by:
\begin{equation}\label{eq:lift}
	\text{Lift}(x) := \frac{\text{share of observed investment uptake in top $x$ customers}}{\text{baseline investment uptake}}\times,
\end{equation}
where the baseline investment uptake is the proportion of all Premier customers that took an investment in the given time period.
The lift values tell us how much more likely the top $x\%$ as ranked by the CLV model are to actually take up an investment in the given time period.
Reading off from the lift curve of \cref{fig:lift}, we see $\text{Lift}(40) = 1.8\times, \text{Lift}(20) = 2.4\times$ and $\text{Lift}(10) = 3.2\times$, meaning, for example, that the top 10\% of predicted investment propensity Premier customers  are around 3.2 times more likely to actually take up an investment in the next 12 months, than if we were to select $10\%$ of the Premier customer base at random. 

In the ongoing investment customer contact campaign, these propensity models are being used to select target populations by ranking the non-investment holding customers according to their predicted investment propensity.
The lift values computed on the test set provide confidence that the top 10\%, 20\% and 40\% of predicted investment propensity Premier customers are indeed more likely to invest, allowing for more targeted use of marketing resources for those customers.
Similar approaches are also being used to select target populations as part of a range of customer contact campaigns based around specific products, including mortages and investments, as well as other propositions such as signing up to the Premier banking program and Savings Account balance building campaigns.

\section{Summary and conclusion\label{sec:conclusion}}

In this paper, we have developed a CLV modelling methodology that is applicable to industries involving long-lasting contractual products and we have demonstrated an application of this model in the retail banking industry. 
Our methodology built on the current state-of-the-art through the use of machine learning and a simulator, which enabled predictions beyond data history limitations. 
The careful choice of forced splits in the segmentation model, informed both by marketing campaign requirements and data analysis, maintained a CLV model that is useful in supporting product acquisition and nurturing customer relationships. 
Through applying the CLV model in the retail banking industry, we validated our approach on a test set for one- and two-year predictions and observed improved performances over the literature benchmark, where we saw a percentage improvement of 43\% of one-year out-of-time median absolute error over the baseline (\Cref{table:results_oneyear}). 
Moreover, the model can be used to support marketing campaigns through derived propensity models, and we saw how our model is able to extract customers with high propensity to take up specific products in testing (\Cref{fig:lift}).
In particular, the top 10\% of customers ranked by their investment propensity were 3.2 times more likely than a customer chosen at random to take up an investment product in the following year.

Although the application of our CLV model in the retail banking industry demonstrated impressive results, there are some limitations and avenues for future improvements, which we now discuss. 
The most notable is the lack of validation for longer-term predictions, which is caused in any use case by finite limitations on data history. 
In our particular use case the data history was limited to two years.
Whilst the primary usage of the CLV model outputs used only the one-year predictions, longer-term predictions will be used to drive strategic decisioning. 
Hence we should have confidence in the accuracy of longer-term CLV predictions and so validating predictions that go beyond data history is a necessary future work. 
It is worth testing alternative machine learning models, such as deep neural networks, to see if performance can be further improved for both one-year and two-year predictions.
Customer embeddings can also be generated by deep learning, as done by \citet{chamberlain2017customer}, in order to enrich the training data.
Reinforcement learning may be suitable for managing customer relationships and this could be an interesting future research direction.

\subsubsection*{Acknowledgements.} This project was completed with the support of teams comprising numerous data scientists, data engineers, marketing specialists and project managers. The authors would like to acknowledge the guidance of Zachery Anderson and Graham Smith of NatWest Group. In addition, we would like to acknowledge the efforts across the NatWest Data \& Analytics, Wealth and Retail teams alongside the external support of QuantumBlack, AI by McKinsey, towards the development of the model.


\bibliographystyle{plainnat}


\end{document}